\documentclass[sigconf]{acmart}
\usepackage{makecell}
\AtBeginDocument{%
  }

\copyrightyear{2023}
\acmYear{2023}
\setcopyright{acmlicensed}\acmConference[ICMI '23]{INTERNATIONAL CONFERENCE ON MULTIMODAL INTERACTION}{October 9--13, 2023}{Paris, France}
\acmBooktitle{INTERNATIONAL CONFERENCE ON MULTIMODAL INTERACTION (ICMI '23), October 9--13, 2023, Paris, France}
\acmPrice{15.00}
\acmDOI{10.1145/3577190.3614164}
\acmISBN{979-8-4007-0055-2/23/10}

\begin{document}

\title{Do I Have Your Attention: A Large Scale Engagement Prediction Dataset and Baselines}

\author{Monisha Singh}
\affiliation{%
 \institution{Indian Institute of Technology}
 \city{Ropar}
 \state{Punjab}
 \country{India}}
\email{monisha.21csz0023@iitrpr.ac.in}
\orcid{0000-0003-0373-8169}

\author{Ximi Hoque}
\affiliation{%
 \institution{Indian Institute of Technology}
 \city{Ropar}
 \state{Punjab}
 \country{India}}
\email{2021aim1015@iitrpr.ac.in}
\orcid{0000-0001-7430-6270}

\author{Donghuo Zeng}
\affiliation{%
 \institution{KDDI Research, Inc.}
 \city{Fujimino-shi}
 \state{Saitama}
 \country{Japan}}
\email{xdo-zen@kddi.com}
\orcid{0000-0002-6425-6270}

\author{Yanan Wang}
\affiliation{%
 \institution{KDDI Research, Inc.}
 \city{Fujimino-shi}
 \state{Saitama}
 \country{Japan}}
\email{wa-yanan@kddi-research.jp}
\orcid{0000-0001-6562-0487}

\author{Kazushi Ikeda}
\affiliation{%
 \institution{KDDI Research, Inc.}
 \city{Fujimino-shi}
 \state{Saitama}
 \country{Japan}}
\email{kz-ikeda@kddi-research.jp}
\orcid{0009-0000-9563-760X}

\author{Abhinav Dhall}
\affiliation{%
 \institution{Indian Institute of Technology}
 \city{Ropar}
 \state{Punjab}
 \country{India}}
\affiliation{%
 \institution{Monash University}
 \city{Melbourne}
 \country{Australia}}
\email{abhinav@iitrpr.ac.in}
\orcid{0000-0002-2230-1440}

\renewcommand{\shortauthors}{Singh et al.}

\begin{abstract}
  The degree of concentration, enthusiasm, optimism, and passion displayed by individual(s) while interacting with a machine is referred to as `user engagement'. Engagement comprises of behavioral, cognitive, and affect related cues~\cite{https://doi.org/10.1002/asi.21229}. To create engagement prediction systems that can work in real-world conditions, it is quintessential to learn from rich, diverse datasets. To this end, a large scale multi-faceted engagement in the wild dataset \emph{EngageNet} is proposed. 31 hours duration data of 127 participants representing different illumination conditions are recorded. Thorough experiments are performed exploring the applicability of different features, action units, eye gaze, head pose, and MARLIN. Data from user interactions (question-answer) are analyzed to understand the relationship between effective learning and user engagement. To further validate the rich nature of the dataset, evaluation is also performed on the EngageWild dataset. The experiments show the usefulness of the proposed dataset. The code, models, and dataset link are publicly available at https://github.com/engagenet/engagenet\_baselines.
\end{abstract}




\maketitle

\section{Introduction}
COVID-19 has accelerated the adoption of online meetings, education classes, and business activities. From online classes to online product launches, virtual meetings have been adopted and accepted rapidly. Although these online meetings have numerous advantages, such as accessibility and affordability, it suffers from a few challenges in the form of engagement. Let's take the example of a classroom; for an instructor, it is easier to gauge a student's interest in the didactic content taught in a classroom-based setting by looking for signs like head nodding and yawning. However, observing such a change in behavior in online classroom settings is challenging. It is necessary to develop automatic mechanisms for detecting student engagement to increase student participation and decrease academic ennui. Teachers and instructors can adjust their content and methods to cater to the interests of their students with the use of engagement detection technologies.

\begin{figure}[t]
\centerline{\includegraphics[scale=0.22]{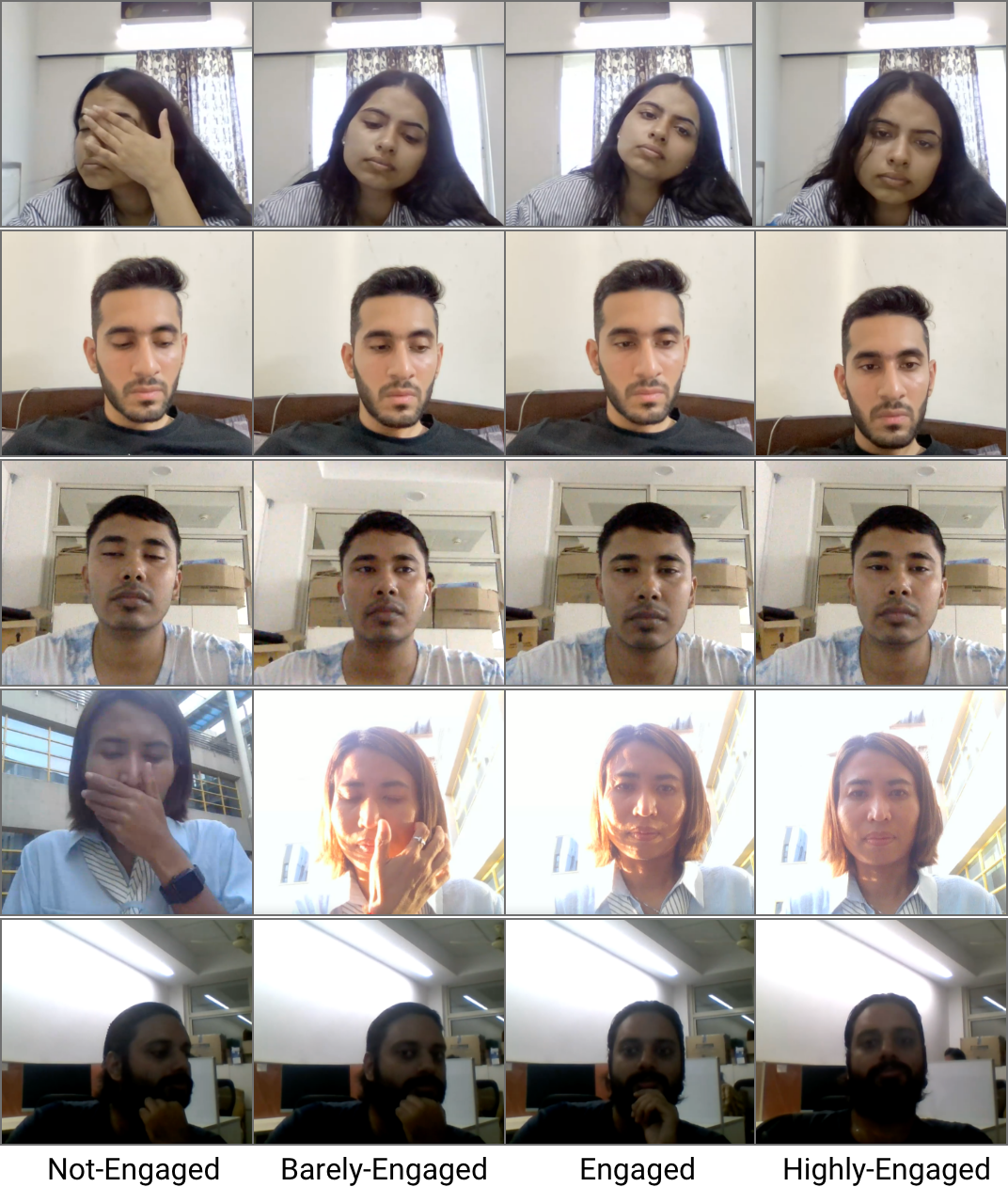}}
\caption{Frames from the proposed \emph{EngageNet} dataset. The engagement categories are mentioned at the bottom. Notice the `in-the-wild' aspect of the data.} 
\label{fig}
\end{figure}

The necessity of measuring engagement is not limited to education settings but also in UX research during scenarios such as A/B testing and even in human-robot interaction. From the perspective of social robots, how can a robot understand if the user is engaged in an activity and an interrupt from the robot may disturb the user? Similarly, if a user is not engaged in the conversation, the robot can stop speaking. Measuring user engagement across various domains is crucial. Additional examples of applications where it's crucial to gauge user engagement are presented as follows: It's critical to create engaging and engrossing games in the gaming industry. Information on user interaction can be utilized to pinpoint the game's most captivating elements and modify its design accordingly. In digital marketing, measuring user engagement is extremely crucial. By examining user interaction on websites and social media platforms, marketers can modify their advertising strategies. Measuring user engagement is also crucial in digital health applications. Healthcare professionals can assess how well patients are following their treatment plan and modify their treatment as necessary. For such circumstances, it is important to predict user engagement.

Several aspects of user engagement have been canopied in the literature. Three predominantly identified characteristics, behavioral, cognitive, and affective, consequently bear a considerable impact on user engagement \cite{kaur2018prediction}. Behavioral engagement is the user's visible participation in their learning; head pose, eye gaze, and facial expression are some of the visual cues of behavioral engagement  \cite{kaur2018prediction}. Cognitive engagement is the degree to which students are prepared and competent to tackle their learning tasks. Affective engagement refers to how the user feels in a particular situation (e.g., amused, joyful, or proud, etc.). Designing and creating student engagement technologies requires access to statistics that capture the three aspects of engagement. A few publicly accessible datasets (DAiSEE~\cite{gupta2016daisee}, HBCU~\cite{6786307}, EngageWild~\cite{kaur2018prediction}) that include behavioral, cognitive, or affective components of engagement partially fill this demand. However, there are not any large sized publicly available datasets that are currently accessible to the general public that records multiple facets of engagement. By creating a large dataset, which captures both the behavioral and cognitive aspects of engagement, we try to address this limitation in our paper. We refer to the proposed dataset as \emph{EngageNet}. The dataset is generated by designing a study in which participants watch three stimuli videos and answer questions based on the stimuli videos over a web-based interactive platform. A personality-based questionnaire, three video stimuli, and three questionnaires based on the stimuli make up the core of our study design. The main contributions of the paper are described below:

\begin{itemize}
  \item Large scale dataset EngageNet with 31 hours of data of 127 subjects, and over 11.3K 10-sec clips is presented.
  \item  To capture multiple facets of engagement, we included behavioral as well as cognitive components of engagement in dataset recording.
  \item To establish baselines for evaluating behavioral engagement in multimodal setups, we conducted extensive experiments.
  \item In the video stimuli, we introduced digital teachers (virtual avatars) to study effect of digital avatars on engagement in future.
\end{itemize}

For contributing to the research in the community, the code, evaluation protocols, features, and data have been made publicly available.

The remainder of the paper is structured as follows:  Section 2 provides a quick overview of the user engagement prediction methodologies and existing engagement datasets. We describe the proposed dataset in Section 3, which includes the study design, data collection, pre-processing, annotation, data split, and statistics. In Section 4, we go over experiments and their results. In Section 5, we cover cognitive engagement analysis. In section 6, the limitations of the proposed dataset. We conclude and discuss the future directions in Section 7.

\begin{table*}[t]
\caption{Comparison of engagement labelled datasets.}
\begin{center}
\resizebox{\linewidth
}{!}{%
\begin{tabular}{ccccccc}
\hline
\textbf{Dataset} & \textbf{Setting} & \textbf{\#Participants} & \textbf{Annotators} & \textbf{Duration} & \textbf{Total Hours} & \textbf{\makecell{Level of Abstraction-\\Combination}}\\
\hline\hline
DAiSEE~\cite{gupta2016daisee} & Virtual learning, lecture, non-interactive, in the wild & 112 & \makecell{Observers\\ (10 Non-Experts)} & 10 sec & 25 
& \makecell{Affective\\dimension}\\
\hline
HBCU~\cite{6786307} & Cognitive Skills Training~\cite{inproceedings}, interactive & 34 & \makecell{Observers\\ (10 sec-7 labelers\\
60 sec-2 labelers)} & 10, 60 sec & 25.5
& \makecell{Behavioral,\\Cognitive\\dimension}\\
\hline
EngageWild~\cite{kaur2018prediction} & Computer based, lecture, non interactive, in the wild
& 78 & \makecell{Observers\\ (5 experts)} & 5 min & 16.5
& \makecell{Behavioral\\dimension}\\ 
\hline

\makecell{EngageNet\\ (Ours)} & \makecell{Computer based, talk, interactive, in the wild, \\digital teacher avatar} & 127 & \makecell{Observers\\ (3 experts)} & 10 sec
& 31 & \makecell{Behavioral,\\Cognitive\\dimension}\\ 
\hline
\end{tabular}%
}
\label{tab1}
\end{center}
\end{table*}

\section{Related Work}
Engagement prediction is a data driven problem, and there are a handful of datasets available for public use \cite{kaur2018prediction, gupta2016daisee}. There have been several initiatives to address the requirement for freely accessible datasets for training engagement detection models. In this section, we first discuss the standard engagement labelled datasets. 

The DAiSEE dataset, created by Gupta et al.~\cite{gupta2016daisee}, includes 112 subjects, of which 80 are males and 32 are females. Videos in the dataset were recorded under three different lighting conditions—light, dark, and neutral—in unconstrained locations, such as dormitories, labs, and libraries. The video of the participants watching a video lecture was recorded by a webcam that was mounted on a Computer. The 10 second long videos in the dataset have been annotated by 10 unskilled crowdsourcing workers into four different levels: engaged, bored, confused, and frustrated. 

The HBCU dataset~\cite{6786307} was gathered from undergraduate students who took part in a "Cognitive Skills Training Project" that ran between 2010 and 2011. The experimental data from 34 subjects(9 males and 25 females) were split into two groups:  (a) 26 subjects who participated in the Cognitive Skills Training study in the spring of 2011. (b) 8 subjects participated in the Cognitive Skills Training project in the summer of 2011 at the University of California (UC). The test subjects either sat by themselves or with the experimenter while the cognitive skills training software was being played in a private space. A group of labelers was assembled from HBCU and UC computer science, cognitive science, and psychology departments, including both undergraduate and graduate students.

Kaur et al.~\cite{kaur2018prediction} provided EngageWild dataset for the creation of their proposed method for detecting student engagement. Videos of 5-minute duration were captured from 78 individuals — 25 females and 53 males. To gather data, they employed a variety of settings, including computer labs, dormitories, the outdoors, etc. A group of five annotators rated the subjects' level of participation in the videos. \emph{Not Engaged}, \emph{Barely Engaged}, \emph{Engaged}, and \emph{Highly Engaged} were the four possible engagement levels that were used to annotate the data. They used weighted Cohen's K with quadratic weights as the performance parameter to evaluate the reliability of annotators. Additionally, they have successfully addressed the issue of subjective bias and low annotator agreement. A brief comparison of the existing datasets is presented in Table~\ref{tab1}.

Approaches for detecting engagement based on computer vision gather data from gestures, eye gaze, head movements, and facial expressions. This gathered data is then fed into the machine or deep learning models. Here we will go over a couple of the methods that have been suggested for engagement detection. One of the seminal works in this direction is by Whitehill et al.\cite{6786307}. They employed binary classifiers for the classification task. Three classifier combinations have been compared: GentleBoost with Box Filter features, Support vector machines with Gabor features, and Multinomial logistic regression with expression outputs from the Computer Expression Recognition Toolbox~\cite{littlewort2011computer}. In the final stage, a regressor was fed the outputs of the binary classifiers to detect engagement automatically. To predict the subject's engagement level, Kaur et al.~\cite{kaur2018prediction} proposed a segment-based random forest model and LSTM-based deep sequence network baselines. For the classification job, Thomas et al.~\cite{10.1145/3139513.3139514} used machine learning methods, including Support Vector Machine (SVM) and Logistic Regression (LR). In a different paper, Thomas et al.~\cite{10.1145/3242969.3264984} provided a system for measuring student engagement levels using the Temporal Convolutional Network (TCN). For video classification, Gupta et al.~\cite{gupta2016daisee} used a 3D CNN model for full training and employed a long-term Recurrent Convolutional Network for unifying visual and sequence learning. On the DAiSEE dataset, Ali et al.~\cite{abedi2021affect} accurately identified disengaged videos and achieved better classification accuracy by employing a variety of temporal(LSTM and TCN) and Non-temporal(fully connected neural network, SVM, and RF) models.

Between our EngageNet dataset and past efforts, we discovered significant variances. First off, very few of the earlier studies address multiple aspects of engagement. Our dataset, in turn, records engagement's behavioral and cognitive facets. Second, to the best of our knowledge, our dataset is the largest dataset captured in the wild in varied illumination settings. Third, in addition to the labels that expert labelers annotated, we also collected the respondents' retrospective self-labels. Finally, we have obtained personality data from participants that can help us understand how engagement and personality relate to one another in future works.

\section{EngageNet Dataset}

\subsection{Study Design}
A web based platform was designed for the data collection study. The participants were between the age range of 18 to 37 years. The usage of a computer or laptop with a quality webcam and a steady internet connection was advised to the participants. The participant could use the web based interface anywhere they felt comfortable. This enabled collection of data in diverse environments. We acquired demographic data including age and biological sex. Participants' consent was also requested for the scope of usage of the data. Participants were also required to complete a pre-study personality-based questionnaire. 

To account for the subjective bias of the annotators, we presented three stimuli videos that were randomly shuffled at the start of each session. The average length of the study session was about 25 to 30 minutes, and each video stimulus was about 5 minutes long. To gauge the subject's level of cognitive engagement, a brief questionnaire based on the content of each video stimulus was presented afterward. For each stimulus, the questionnaire contained two questions of difficulty easy to moderate. The participants were asked to self-identify their perceived level of engagement after completing the brief questionnaire. The participants were also questioned if they had previously watched the video, whether they found the content of the video interesting, and would they recommend it to others.

\subsection{Data Collection, Pre-processing and Annotation}
\subsubsection{Data Collection}
\begin{figure}[t]
\centerline{\frame{\includegraphics[scale=0.3]{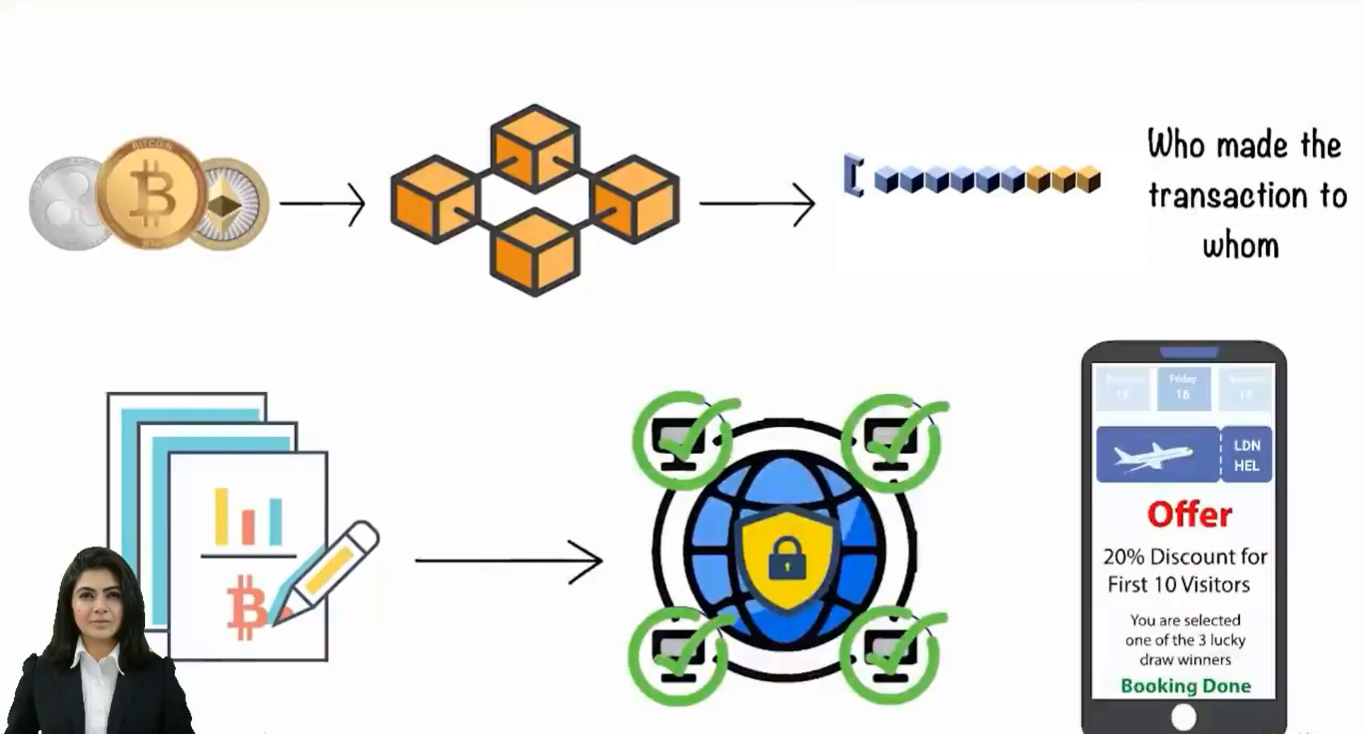}}} 
\caption{A frame from one of the stimuli video used in data collection. Note the digital avatar on the bottom left.} 
\label{StimuliPic}
\end{figure}

In this section, we discuss the study's data recording paradigm. Broadly two sets of information are captured during the experiment: (i) capturing the participants on camera when they watch the stimulus video and (ii) a brief questionnaire is answered by the participant based on the content of the viewed video stimulus. We used three educational videos as stimulus: (i) \emph{Schrödinger's cat: A thought experiment in quantum mechanics} (ii) \emph{What is cryptocurrency?} (iii) \emph{Where did English come from?}. To appeal to a wider audience it was important to select videos with different themes. The audio part was orated by a digital avatar added to the videos. The reason to add a digital avatar (teacher) is to check for the engagement in videos with a digital teacher and without it. Note these experiments will be a future work. The digital teacher has been added using \footnote{The Artiste AI Studio - www.kroop.ai}. A screenshot of the stimuli is shown in Figure \ref{StimuliPic}. 

There are 127 participants in the dataset (83 males and 44 females). The participants are between the ages of 18 and 37 years. A total of 474 videos, each lasting 5 to 10 minutes depending on the time the subjects took to answer the questions, were gathered. The dataset is gathered in an unconstrained environment at various sites, including computer labs, dorm rooms, open spaces, etc. The videos were recorded in at least 640x480 pixels resolution. The diverse dataset includes videos with background noise. Performance of the system is impacted by noise and a lack of audio, thus we only concentrated on video.

\subsubsection{Data Pre-processing}

Overall for the EngageNet dataset, the video duration against stimuli and question response varied between 5 to 10 minutes. The videos were further divided into 10 seconds duration. This was performed as the engagement level varies over time in long videos \cite{10.1145/3382507.3417973}. Moreover, Whitehill et al.\cite{6786307} used 10 sec videos and noted that 10 sec videos were intuitive from labeling perspective. The final data contains videos of 10 seconds in duration. 

\subsubsection{Annotation}
The participants in the videos were assigned to different engagement classes by a group of three annotators. We employed the four-engagement class definitions that Kaur et al.~\cite{kaur2018prediction} stated. The following are the engagement classes: \emph{"Not Engaged"} denotes a wholly disengaged subject, such as one who frequently glances away from the screen and appears disinterested. \emph{"Barely Engaged"} refers to being only minimally attentive, like when the individual fidgets restlessly in the chair or hardly opens his or her eyes. The term \emph{"Engaged"} refers to participants who are interested in the content, such as when they interact with the video and appear to like it. \emph{"Highly Engaged"} refers to a subject who appears to be entirely attentive and glued to the screen. We have used majority voting for assigning the final labels. In scenarios where all three annotators assigned different labels, we used an academic expert. Additionally, we used weighted Cohen's K~\cite{cohen1968weighted} with quadratic weights as the performance parameter to evaluate the reliability of annotators. Weighted Kappa score was found to be 0.73 between first and second labeler, 0.79 between first and third labeler, and 0.77 between second and third labeler. A Kappa value between 0.61 and 0.80 is regarded as substantial~\cite{article}.

\subsection{Data Split and Statistics} 
We created subject independent data split leading to 90 participants in \emph{Train}, 11 participants in \emph{Validation}, and 26 participants in \emph{Test}. 
Total videos for \emph{Train}, \emph{Validation}, and \emph{Test} sets were 7983, 1071, and 2257, respectively. The study included 127 participants, of which 65.35\% were male, and 34.65\% were female. The participants were between the ages of 18 and 37 years. The dataset consists of approximately 11.3K clips lasting 10 seconds, or $\sim$31 hours of data. The class distribution of labels is as follows: \emph{Highly Engaged} (48.57\%), \emph{Engaged} (20.97\%), \emph{Barely Engaged} (12.3\%), and \emph{Not Engaged} (18.16\%). 

\begin{table*}[htbp]
\begin{center}
\resizebox{\linewidth
}{!}{%
\begin{tabular}{|l|l|l|}
\hline
\textbf{Model} & \textbf{Layer} & \textbf{Hyperparameters} \\ 
\hline\hline
\textbf{LSTM} & \makecell[l]{LSTM Layer\\ Dropout Layer\\ Fully Connected Layer} & \makecell[l]{Units: [256, 128]\\ Rate: 0.5\\Units: [128, 64, 4], Activation: [ReLU, ReLU, Softmax]}\\
\hline
\textbf{CNN-LSTM} & \makecell[l]{Convolutional Layer\\MaxPooling Layer\\LSTM Layer\\Dropout Layer\\Fully Connected Layer} &
\makecell[l]{Filters: [128, 64], Kernel Size: [8, 8], Activation: [ReLU, ReLU]\\Pool Size: 4\\Units: [128, 64]\\Rate: 0.5\\Units: 4, Activation: Softmax}\\
\hline
\textbf{TCN} & \makecell[l]{Temporal Convolutional Layer\\Dropout Layer\\ Fully Connected Layer} &
\makecell[l]{Filters: [24, 24], Dilations: [[1, 2, 4, 8], [1, 2, 4, 8]]\\Rate: 0.5\\Units: [128, 4], Activation: [ReLU, Softmax]}\\
\hline
\textbf{Transformer} & \makecell[l]{Transformer Encoder Layer\\ Dropout Layer\\Fully Connected Layer} & \makecell[l]{Number of heads: 8, Head size: 256, Number of blocks: 4\\Rate: 0.3\\Units: 128}\\
\hline
\textbf{Transformer Fusion} & \makecell[l]{Transformer Encoder Layer 1\\Transformer Encoder Layer 2\\Dropout Layer\\Fully Connected Layer} & \makecell[l]{Number of heads: 8, Head size: 256, Number of blocks: 4\\Number of heads: 8, Head size: 256, Number of blocks: 4\\Rate: 0.3\\Units: 128}\\
\hline
\end{tabular}%
}
\caption{Hyperparameter details of the different networks used for baseline evaluation.}
\label{tab:hyperparameters}
\end{center}
\end{table*}

\begin{table*}[t]
\begin{center} \label{ComparisonTable}
\resizebox{\linewidth
}{!}{%
\begin{tabular}{|l|c|c|c|c|c|c|c|c|c|c|c|c|c|c|c|}
\hline
\textbf &\multicolumn{5}{c|}{\textbf{LSTM}} & \multicolumn{6}{c|}{\textbf{CNN-LSTM}} & \multicolumn{4}{c|}{\textbf{TCN}}\\
\hline
&\textbf{G}&\textbf{HP}&\textbf{AU} & \textbf{G+HP}&\textbf{G+HP+}&\textbf{G} &
\textbf{HP}&\textbf{AU}&\textbf{G+AU}&\textbf{HP+AU}&\textbf{G+HP+}&\textbf{G}&\textbf{HP}&\textbf{AU}&\textbf{G+HP+}\\
& & &  & &\textbf{AU}& & & & & &\textbf{AU} & & & &\textbf{AU}\\
\hline
\textbf{Val.} & 61.25 & 67.60 & 63.03 & 67.69 & 67.04 & 60.60 & 67.32 & 61.72 & 62.75 & 67.51 & 67.51 & 62.56 & 66.11 & 62.93 & \textbf{67.79} \\
\hline
\textbf{Test} & 59.84 & 62.46 & 60.06 & 62.37 & 61.84 & 58.73 & 58.82 & 59.26 & 61.13 & 64.05 & 65.16 & 59.26 & 59.13 & 58.73 & \textbf{65.60} \\
\hline
\end{tabular}%
}
\caption{Comparison of accuracy(\%) of different feature combinations. Here, G:Eye Gaze, HP:Head Pose, AU:Facial Action Units.}
\label{tab2}
\end{center}
\end{table*}

\begin{figure}[t]
\centerline{\includegraphics[scale=0.6]{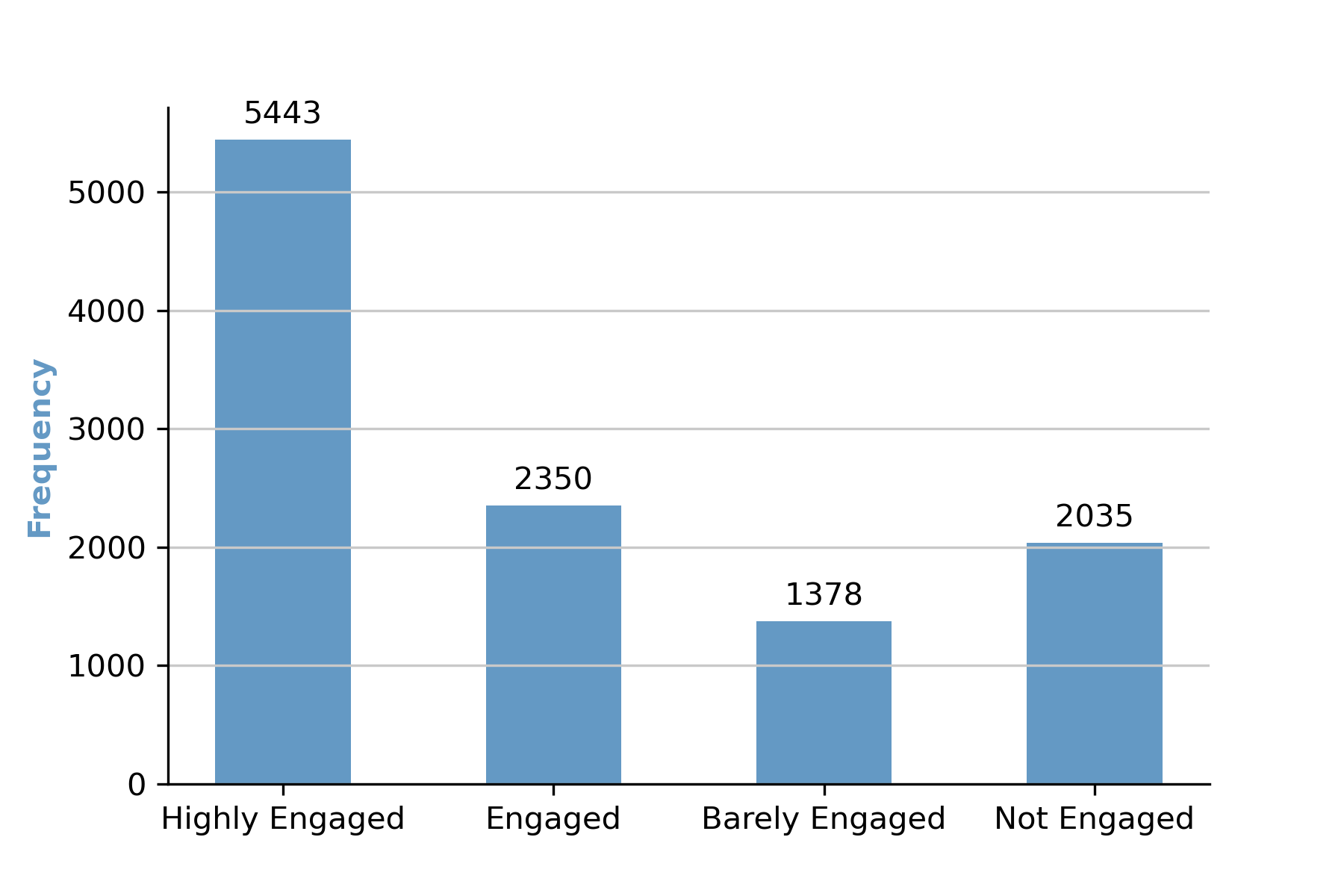}}
\caption{Distribution of Engagement Classes in EngageNet.}
\label{figGraph}
\end{figure}

\section{Experiments}
In this section, we discuss the quantitative analysis on the EngageNet dataset along with baselines. Both hand engineered and representation learning based are features extracted from the facial region in the EngageNet videos. The feature and classifier details are described below. 

\subsection{Feature Extraction} 
Face and facial landmark positions were identified in each frame using the OpenFace toolkit~\cite{10.1109/FG.2018.00019}. For facial landmark detection and tracking, OpenFace 2.0~\cite{10.1109/FG.2018.00019} makes use of the Convolutional Experts Constrained Local Model~\cite{8265507}. Internally, CE-CLM projects face landmarks onto the image using an orthographic camera projection method. Once the landmarks are located, they solve the n point in perspective problem~\cite{10.1109/ICCV.2011.6126266} and precisely estimate the head pose. Constrained Local Neural Field (CLNF) landmark detector~\cite{10.1109/ICCVW.2013.54}, ~\cite{10.1109/ICCV.2015.428} is employed to recognize the pupil, iris, and eyelids in order to estimate eye gaze. A window of frames' worth of features is taken from the subsampled video. The following batch of frames is then extracted to compute features before the window slides. We computed the mean and standard deviation of head location, head rotation, gaze vector and gaze angle of both eyes, facial action units to capture the changing head pose, gaze, and facial expression in each segment. The final dimensions of eye gaze, head pose, and facial AUs features were 16, 12, and 70, respectively, after calculating the mean and standard deviation. 

In addition to the features extracted using OpenFace~\cite{10.1109/FG.2018.00019}, we also extracted Masked Autoencoder for facial video Representation LearnINg (MARLIN)~\cite{Cai_2023_CVPR} features. By learning to encode spatio-temporal changes in the face, MARLIN can capture the dynamics of facial movements and expressiveness, which can be used to extract generic and transferable facial features. Unlabelled videos available on the internet are used to train a masked autoencoder framework. Face segmentation based weights are used for different facial regions masking. By using this masking approach, MARLIN focuses on learning and encoding the changes and patterns occurring in different facial regions, allowing it to capture fine-grained facial details. Linear probing and fine-tuning  are used by MARLIN to demonstrate the robustness, generalizability, and transferability of the learned feature for facial task specific downstream adaptation. The learned features are assessed on numerous facial tasks in the linear probing step without any extra training or adaptation. This helps assess the generic and transferable nature of the features learned by MARLIN. The learnt characteristics are tailored to particular facial tasks in fine-tuning, enabling the framework to show that it can be utilized successfully in subsequent applications. We extracted 1024 dimension features by training ViT-large~\cite{dosovitskiy2021an} encoders on YTF~\cite{10.1109/CVPR.2011.5995566} dataset. The YTF dataset (YouTube Faces) is a publicly available dataset consisting of 3,425 videos of 1,595 different people collected from YouTube. The videos show people in real-life conditions, with variations in pose, illumination, expression, age, and resolution. MARLIN code and models are publicly available \footnote{https://github.com/ControlNet/MARLIN}.

\subsection{Engagement Prediction Baselines}
To predict the engagement, we experimented with different classifiers described in this section. The hyperparameter details of the baseline models are provided in table ~\ref{tab:hyperparameters}.

\subsubsection{Long Short-Term Memory Network}
Deep networks built on LSTMs~\cite{10.1162/neco.1997.9.8.1735} were chosen because they are effective at identifying long-term dependencies in sequential data. The network is made up of two LSTM layers, which are followed by a Dropout layer to stop the model from overfitting and a Flatten layer to get a single activation vector for all segments of the video. The resulting feature vector is fed into three fully connected layers, two of which have ReLU activation, while the third layer has softmax activation. Empirically, we observed 10 cells (segments) as optimum setting.

\subsubsection{CNN-LSTM}
The CNN-LSTM~\cite{DBLP:journals/corr/DonahueHGRVSD14} based model was chosen as CNNs have demonstrated high accuracy in several computer vision tasks and LSTMs are widely recognized for their learning dependencies between time steps in time series and sequential data. Two convolutional layers, MaxPooling, TimeDistributed Flatten, two successive LSTM layers, Dropout, Flatten, and ultimately a fully connected layer with softmax activation make up the model.

\subsubsection{Temporal Convolutional Network}

The TCN~\cite{8099596} based network was chosen since it has proven to be more effective than other sequential models like LSTMs and can capture long-term dependencies. To produce a single vector of activations, the TCN-based deep model comprises of two TCN layers placed one after the other, followed by a Dropout layer and a Flatten layer. The resultant vector is then passed into a series of fully connected layers, the last layer of which is activated using softmax.

\subsubsection{Transformer}
Recent works \cite{10.1145/3505244} in computer vision have witnessed substantial increase in performance due to the Transformer architecture~\cite{10.5555/3295222.3295349}. We experimented with a transformer architecture with positional encoding, an attention layer stacked above and a MLP to predict engagement. The tokens were created by dividing the 10 seconds video samples into 20 segments of uniform duration. Each segment's final feature is its mean and standard deviation computed from the features belonging to a segment. 20 tokens was chosen after empirical evaluation.

\begin{table}[t]
\begin{center}
\resizebox{\columnwidth
}{!}{%
\begin{tabular}{|l|c|c|c|}
\hline
\textbf{Features}&\textbf{Validation}&\textbf{Test}\\
\hline
\textbf{Gaze} & 55.45 & 54.87  \\
\hline
\textbf{Gaze + Head Pose} & 64.45 & 62.19  \\
\hline
\textbf{Gaze + Head Pose + AU} & \textbf{69.10} & \textbf{67.61}  \\
\hline

\end{tabular}%
}
\caption{Accuracy comparison (\%) of Transformer model on different combination of different features.}
\label{tab3}
\end{center}
\end{table}

\subsubsection{Fusion Model}
The goal of a fusion model is to leverage the strengths of each input source or model and combine them to produce a more accurate or robust prediction. We chose the following fusion strategy, representations from trained OpenFace and MARLIN based features are concatenated with a MLP.

\begin{table}[t]
\begin{center}
\resizebox{7cm}{!}{%
\begin{tabular}{|l|c|c|}
\hline
\multicolumn{2}{|c|}{\textbf{Transformer Fusion Model}}\\
\hline
\textbf {} &\textbf{G+HP+AU+MARLIN}\\
\hline
\textbf{Validation} & 68.49\\
\hline
\textbf{Test} & 66.5\\ 
\hline
\textbf{Train+Validation} & \textbf{69.29}\\ 
\hline
\end{tabular}%
}
\caption{Comparison of accuracies (\%) of fusion model. Here, G:Eye Gaze, HP:Head Pose, AU:Facial Action Units.}
\label{tab4}
\end{center}
\end{table}

\subsection{Baseline Results}
The following experiments were performed using the extracted features and proposed deep learning baseline models.
\subsubsection{LSTM, CNN-LSTM, and TCN Baseline Evaluations}
Eye gaze, head position, and facial AUs features were input to LSTM, CNN-LSTM, and TCN models, separately. Later, the LSTM model was fed as input a combination of gaze and head pose features, and the CNN-LSTM was given as input first a combination of gaze and facial AUs, and then a combination of head pose and facial AUs. Finally, the three models were fed as input a combination of gaze, head pose, and facial AUs. The outcome of this experiment is shown in Table ~\ref{tab2}.
We observed that multiple features (head pose+gaze+facial AU) outperformed individual features on the \emph{Test} set. Particularly, the combination of eye gaze, head pose, and facial AUs features performs remarkably well in predicting engagement, which is suggestive of the fact that gaze, head pose, and facial expression were all taken into account when annotating the video clips.

\subsubsection{Transformer Baseline Evaluations}
The transformer model was first evaluated with the gaze and followed by a combination of gaze and head pose, gaze, head pose, and facial AU features. Table ~\ref{tab3} shows the outcome of this experiment. It is observed that the fusion of features as tokens performs better than gaze only and gaze + head pose features. We also evaluated the performance of MARLIN features with Transformers architecture and it achieved 65.2\% on the \emph{Test} set.

\subsubsection{Fusion models' Baseline Evaluations}
The performance of the transformer baseline model on the validation and test set was significantly better than the other baseline models; so, we used the Transformer baseline model to build the fusion model for the classification task. Also from Table~\ref{tab2} and Table~\ref{tab3} it is evident that the combination of eye gaze, head pose, and facial AU features performed remarkably well in predicting engagement; therefore, we input these features along with MARLIN features to get better results. The fusion architecture is based on concatenating the fully connected layers' features from Transformers learnt on OpenFace based features and MARLIN features. Later a MLP is added to the concatenated FC features. Table~\ref{tab4} shows the results of these experiments. We observed that the fusion model performed considerably well when MARLIN features were employed along with gaze, head pose, and action unit features. Transformer models typically require a large amount of data to achieve good performance. This could be a reason why the fusion doesn't yield better performance as compared to individual experiments. We also learnt a model by training on \emph{Train} and \emph{Val} sets together and observe an increase in performance.

\begin{table}[t]
\begin{center}
\begin{tabular}{|l|c|c|c|}
\hline
\textbf{Train Dataset}&\textbf{Test Dataset}&\textbf{Test MSE}\\
\hline
EngageWild~\cite{kaur2018prediction} & EngageWild  & 0.082  \\
\hline
EngageNet (ours) & EngageWild & 0.162 \\
\hline
\end{tabular}
\caption{Cross dataset learning performance analysis. Note MSE is used for this following the labels of the EmotiW dataset. For detail refer to Section \ref{EmotiWAnalysis}.}
\label{tab5}
\end{center} 
\end{table}

\begin{figure}[b] 
\centerline{\includegraphics[scale=0.6]{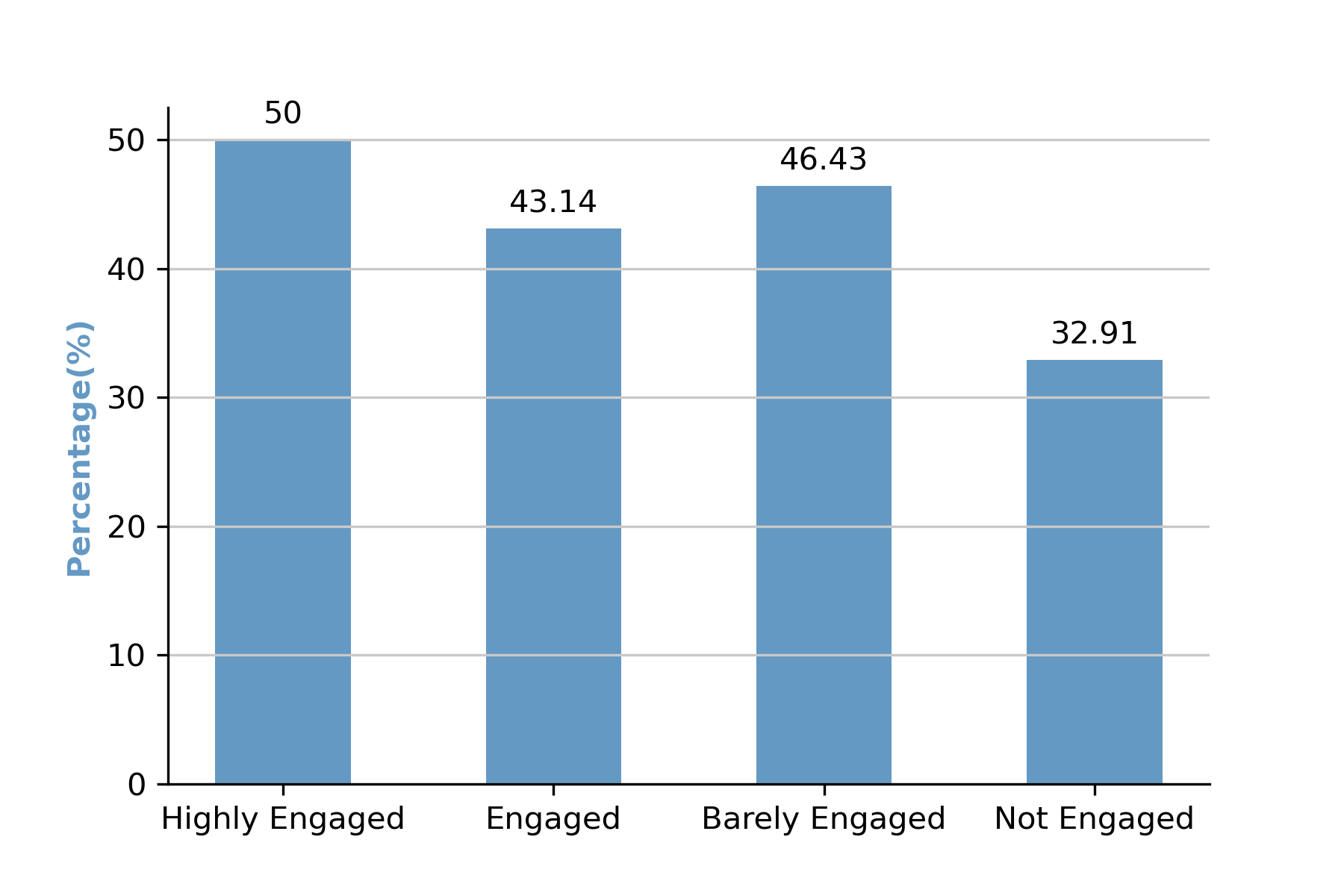}}
\caption{Engagement Labels vs percentage of correct responses per label category. Note that \emph{Not Engaged} class has least correct answers. For details refer to Section \ref{QuestionAnswer}.}
\label{figGraph1}
\end{figure}

\subsubsection{Cross Dataset Analysis} \label{EmotiWAnalysis}
For evaluating the effectiveness and diversity of EngageNet, we evaluated the performance of method trained on EngageNet and evaluated on the EngageWild dataset \cite{kaur2018prediction}. For fair comparison, we follow the protocol and label normalisation mentioned in \cite{kaur2018prediction}. The transformer model was trained in a regression setting to compare the performance of our model with the EngageWild dataset\cite{kaur2018prediction} baseline model of EmotiW challenge\cite{10.1145/3382507.3417973}. In the first setting, the transformer network was trained and tested on the EngageWild dataset. In the second setting, the network was trained on EngageNet and tested on the EngageWild dataset. The performance comparison is shown in Table ~\ref{tab5}. EngageNet trained network achieves relatively comparable performance given the EmotiW label range of 0 to 1. The results show the effectiveness of the proposed EngageNet dataset in terms of the quality of features learnt from the `in the wild' data.

\begin{figure}[t] \vspace{-6mm}
\centerline{\includegraphics[scale=0.6]{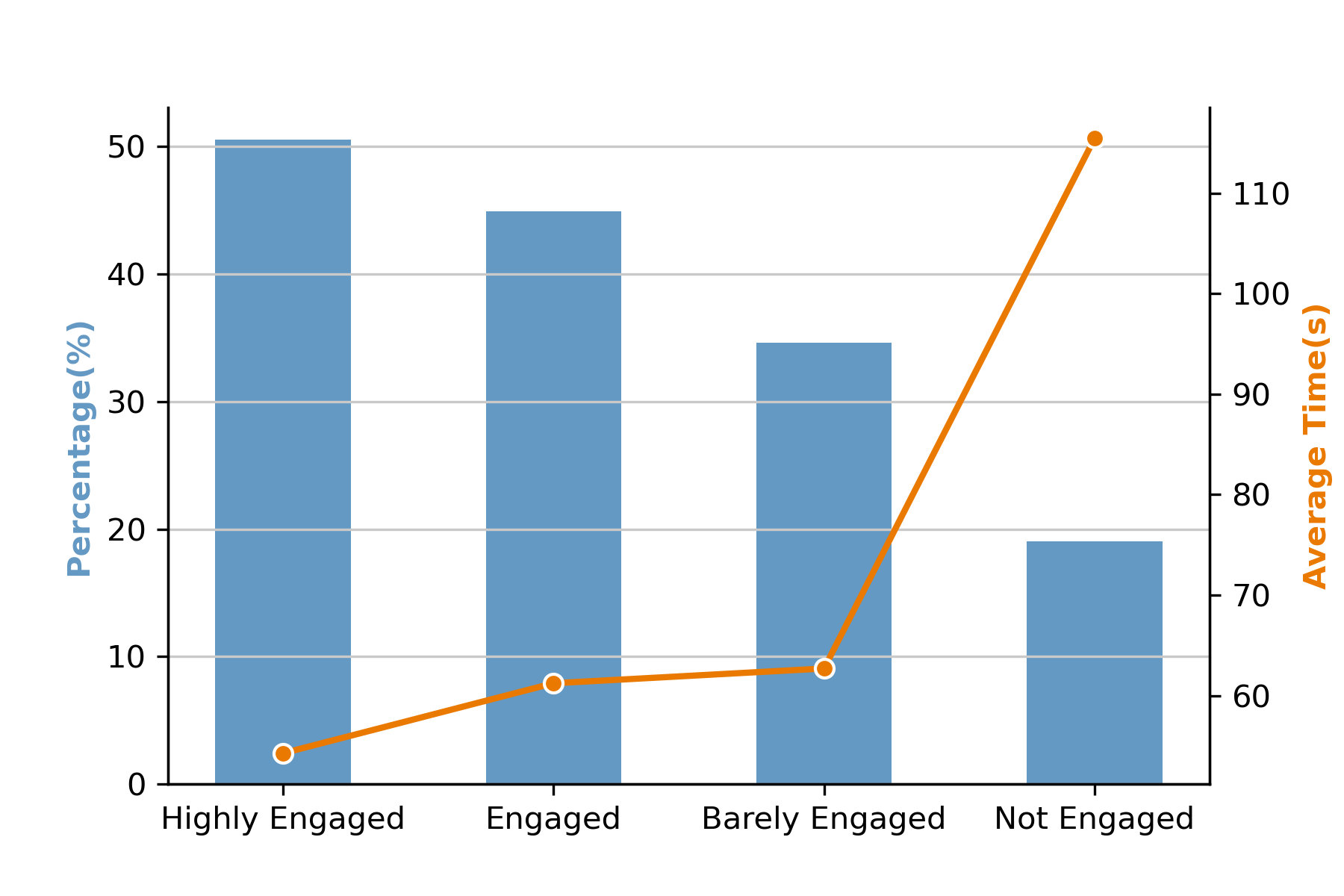}}
\caption{Engagement Classes vs Percentage of correct responses per label category and average response time. Orange color denotes response time in seconds. Note that \emph{Not Engaged} class has least correct answers and longest average response time. The average response time of \emph{Highly Engaged}, and \emph{Engaged} users are lower than other classes. For details refer Section \ref{ChunksAnalysis}. Best viewed in color. }
\label{timeGraph}
\end{figure}

\section{Cognitive Engagement Analysis}
Behavioral, cognitive, and affective are the three most commonly recognized aspects of engagement. Here, we concentrate on engagement's cognitive component. A person's mental effort and involvement in a task or activity is called cognitive engagement. It is a mental state in which a person is totally absorbed in and concentrated on a task, utilizing their cognitive abilities to process and comprehend information and solve problems. Cognitive engagement is considered essential for effective learning and problem-solving. We designed our study in such a way that we can capture the cognitive aspect of engagement. 
In order to assess the users' cognitive engagement in the task at hand, we asked them two questions per stimuli based on the information in the stimuli they watched. We assigned a Boolean value of True only when both answers of the users were correct. We analyzed the performance of users to find out the correlation between the four engagement levels, the user's performance and response time of users.

\subsection{Question Answer Task Analysis} \label{QuestionAnswer}
We analyzed the annotated labels and the performance of participants in the video-based questionnaire to understand the cognitive engagement of the participants. We computed the mode of engagement labels of 10 seconds segments in order to identify the engagement label for the unsegmented videos, which were divided into 10 seconds chunks for annotation. We then calculated the percentage of correct responses (Boolean value True) per label category. The analysis is shown in Figure~\ref{figGraph1}. We found that \emph{Not Engaged} category has the least correct answers and \emph{Highly Engaged} has the highest percentage of correct answers among the engagement classes. This indicates that higher user engagement can lead to better learning. However, the percentage of correct responses for \emph{Barely Engaged} was higher than that of \emph{Engaged}, which contradicts our previous inference that higher user engagement can result in improved learning outcomes. This disparity can be attributed to the fact that the number of instances of \emph{Barely Engaged} is the least compared to other classes. In the future we will be performing detailed statistical analysis to understand the nature of the data and behavior of the users.

\subsection{Engagement Versus Response Time Analysis} \label{ChunksAnalysis}
We extracted those chunks in which the participants were involved in the cognitive task of answering the questions based on the stimuli they watched. To assign the engagement label during the cognitive task, we used the mode of the engagement labels of the cognitive task chunks. We analyzed the percentage of correct responses of the four classes and the average time taken by users to respond. Figure~\ref{timeGraph} depicts this analysis. We observed that the \emph{Highly Engaged} and \emph{Engaged} users during the cognitive task had the maximum percentage of correct responses. Also, \emph{Highly Engaged} and \emph{Engaged} users took less time than \emph{Barely Engaged} and \emph{Not Engaged} users to respond to the questions. This suggests that engagement level can have an impact on task performance and response time. We can infer that the level of cognitive engagement positively correlates with the percentage of correct responses and the speed of responding to the questions. Being highly engaged or engaged during a cognitive task can result in better task performance.

From Figure \ref{timeGraph}, it is also observed that for some cases \emph{Not Engaged} users  took more than 100 seconds to respond to the questions. We investigated these videos and observed that in few cases the users started interacting with their mobile devices. Furthermore, for few users, their individual memory retention and recall ability may be the cause of their delayed reaction. While some individuals may have a natural ability to retain and recall information, others may struggle with these cognitive tasks. 

\section{Limitation}

In this study, we investigated the performance of deep learning models for engagement prediction using the proposed \emph{EngageNet} dataset. The dataset consists of user videos labeled as \emph{Highly Engaged}, \emph{Engaged}, \emph{Barely Engaged}, and \emph{Not Engaged}. There is a class imbalance as evident from the distribution of labels, where the \emph{Highly Engaged} constitutes 48.57\%, while the \emph{Engaged}, \emph{Barely Engaged}, and \emph{Not Engaged} represent 20.97\%, 12.3\%, and 18.16\%, respectively.

The performance of the engagement prediction model could be effected by the class imbalance. The model may get biased and tend to anticipate that the majority of samples will be \emph{Highly Engaged} since the majority class (\emph{Highly Engaged}) greatly dominates the other classes. As a result, while the model's overall accuracy may seem to be high, it may not be able to recognise \emph{Engaged}, \emph{Barely Engaged}, and \emph{Not Engaged} classes with sufficient accuracy. The discovered class disparity prompts inquiries regarding the methodology used for data collection and potential biases in user's response to stimulus. The higher prevalence of \emph{Highly Engaged} in the dataset may be due to users' propensity to be \emph{Highly Engaged} in short laboratory studies. In future research, we propose exploring alternative techniques for handling class imbalance, such as generating synthetic data, or using weighted loss functions. To further increase the generalizability of engagement prediction models, it would be beneficial to employ initiatives to encourage users to express a wider range of engagement. We will also explore synthetic data based strategies for enhancing data points for the lesser represented classes.

\section{Conclusion}
We present a large scale user engagement dataset EngageNet. Both behavioral and cognitive engagement is considered during the dataset creation. We perform extensive baseline analysis with facial features and recent machine learning techniques. The results show the richness and complexity of the data. We observe that highly engaged user answer the largest number of questions correctly as compared with other engagement classes. Additionally, we observed that the time taken by highly engaged and engaged users to answer the questions, as part of the cognitive task, were comparatively lesser than barely engaged and not engaged classes. We also evaluate the data quality by evaluating on another engagement dataset. We believe the dataset, the extensive baselines, and the cognitive analysis will serve as a useful benchmarking resources in the research community.

In the future work, we plan to continue exploring several dimensions related to the user engagement prediction. We observed from the experiments that the fusion of MARLIN with OpenFace based features requires further investigation in the form different fusion architectures. Given the progress of automatic video generation, we will investigate the impact of the presence of digital avatars on user engagement from a multi-lingual automatically created content, perspective. We will analyze stimuli saliency heat maps to gain deeper insights into how users engage with different parts of a stimulus. Another interesting direction is investigating the effect of personality on engagement due to its important implications for the design of personalized user experiences. We will also explore the use of synthetic data to create a more balanced dataset that can improve network performance and generalize better.


\bibliographystyle{ACM-Reference-Format}
\bibliography{sample-base}


\end{document}